%% file: main.tex
\documentclass[twoside,leqno,twocolumn]{article}

\usepackage[letterpaper]{geometry}
\usepackage{appendix}
\usepackage{ltexpprt}
\usepackage{hyperref}
\usepackage{amsmath}
\usepackage{amssymb}
\usepackage{bm}
\usepackage{graphicx}
\usepackage{booktabs}
\usepackage[maxbibnames=99]{biblatex}

\DeclareMathOperator{\Var}{Var}
\DeclareMathOperator{\Cov}{Cov}

\begin{document}

\newcommand\relatedversion{}
\renewcommand\relatedversion{\thanks{The full version of the paper can be accessed at \protect\url{https://arxiv.org/abs/1902.09310}}} 




\title{\Large Metrics for Inter-Dataset Similarity with Example Applications in \\Synthetic Data and Feature Selection Evaluation -- Extended Version}





 \author{Muhammad Rajabinasab
 \and Anton D. Lautrup
 \and Arthur Zimek \and \\  University of Southern Denmark, Odense, Denmark}

\date{}

\maketitle







\begin{abstract} 
\small\baselineskip=9pt Measuring inter-dataset similarity is an important task in machine learning and data mining with various use cases and applications. Existing methods for measuring inter-dataset similarity are computationally expensive, limited, or sensitive to different entities and non-trivial choices for parameters. They also lack a holistic perspective on the entire dataset. In this paper, we propose two novel metrics for measuring inter-dataset similarity. We discuss the mathematical foundation and the theoretical basis of our proposed metrics. We demonstrate the effectiveness of the proposed metrics by investigating two applications in the evaluation of synthetic data and in the evaluation of feature selection methods. The theoretical and empirical studies conducted in this paper illustrate the effectiveness of the proposed metrics.
\end{abstract}

\section{Introduction}
Measuring similarity is an important task in machine learning and data mining with a large number of applications and use cases associated with it. It helps identify patterns, relationships, and structures within data. We often measure the similarity between data points in a dataset for applications such as pattern recognition \cite{julian2012mitchell}, outlier detection \cite{ZimekF18}, information retrieval \cite{nct5hliaoutakis2006information} and recommendation systems \cite{sondur2016similarity}. Similarity can also be measured between features instead of data points to carry out the feature selection task \cite{mitra2002unsupervised}. 

A key research procedure in scientific practice is to measure the similarity between datasets. Consider comparing patient or consumer groups, identifying if experimental results are consistent with each other and with the current paradigm, or checking if a perturbation to an existing dataset caused a significant loss of information. One of the most simple applications of inter-dataset similarity in machine learning and data mining is data augmentation \cite{van2001art}, where similar datasets are identified and used to enhance the training set \cite{hwang2020simex}. Other applications of inter-dataset similarity include synthetic data utility evaluation \cite{Hernandez2022, Dankar2022}, subspace quality assessment \cite{Sammon1969, Kruskal1964}, transfer learning \cite{review_taxonomy_2023}, federated learning \cite{universal_metric_2024}, instance space analysis \cite{smith2023instance}, and dataset selection \cite{ds_sel_10.1007/978-3-031-46994-7_11}.


In this paper, we propose two novel metrics for inter-dataset similarity based on Principal Component Analysis (PCA). PCA \cite{pearson1901liii} is a well-established dimensionality reduction technique which is used widely in different fields of machine learning and data mining \cite{KriKroSchZim12,wiskott2013pca,kandanaarachchi2020,HouKieZim23, aiupbeat2023pca, RAJABINASAB2024126254}. Our metrics, namely \emph{difference in explained variance} and \emph{angle difference}, can be used to measure different aspects of inter-dataset similarity. They can also be utilized to carry out different tasks.

The primary contribution of this paper is the proposed metrics together with an in-depth exploration and discussion of two example applications for the new metrics, namely the evaluation of utility in synthetic data and the evaluation of performance of feature selection algorithms. We also explore their effectiveness in the sanity check of the privacy of synthetic data. Implementation and results are available on \href{
https://github.com/mrajabinasab/Interdataset-Similarity-Metrics}{GitHub}. Even though the applications we present in the paper seem to be specialized, we expect that the functionality of the proposed metrics extends far beyond these use cases to other areas of machine learning and data mining.

The rest of the paper is structured as follows: In~Section \ref{sec_rw}, we briefly review previous work in inter-dataset similarity. In~Section \ref{sec_back}, we discuss the theoretical background of PCA. In Section~\ref{sec_prm}, we introduce and analyze the proposed metrics. In Sections~\ref{sec_synapp} and \ref{sec_fsapp}, we explore and discuss two example applications of the proposed metrics, namely the evaluation of synthetic data and the evaluation of feature selection algorithms in detail. Finally, in Section~\ref{sec_lim}, we discuss limitations of the proposed metric and perspectives for  future work.

\section{Related Work} \label{sec_rw}
Measuring inter-dataset similarity has always been an important task in the field of machine learning and data mining. It has been used both directly \cite{review_taxonomy_2023} and embodied in other metrics and measures \cite{universal_metric_2024}. 
The recent interest in AI-generated content such as images and text has mandated tools for measuring the authenticity hereof. One example of such a measure is the Fréchet Inception Distance \cite{heusel2017gans}, which compares the distribution of one set of images (fakes) to the distribution of another (ground truth). Other metrics such as Kernel Maximum Mean Discrepancy \cite{gretton2012kernel} and Wasserstein distance \cite{panaretos2019statistical}  have also been used to measure a notion of similarity between two datasets. Measures such as Sammon Stress \cite{Sammon1969} and Kruskal Stress \cite{Kruskal1964} are used for evaluating the goodness of fit in the low-dimensional subspaces by assessing distance- and topology preservation respectively. 
In~transfer learning, inter-dataset similarity is used to ensure that the source and target datasets are similar \cite{review_taxonomy_2023}. This makes the transfer of knowledge more effective and improves the performance of downstream tasks. Inter-dataset similarity is also used in federated learning to accurately measure the similarity of the datasets from different silos \cite{universal_metric_2024}. This helps in improving the aggregation process and leads to a better performance for the global model. One other application of inter-dataset similarity is to find the most different dataset for the empirical analysis of an arbitrary machine learning or data mining method \cite{ds_sel_10.1007/978-3-031-46994-7_11} to ensure a robust and comprehensive empirical analysis.

Existing metrics can measure a notion of similarity between two datasets, but they have their own challenges and shortcomings. Fréchet Inception Distance is model-dependent as it relies on a pre-trained inception model. Maximum Mean Discrepancy is computationally expensive and depends on the choice of the kernel which can be non-trivial. Wasserstein distance has a high computational complexity and is sensitive to outliers. Sammon and Kruskal Stress also have a high computational complexity, Sammon Stress is sensitive to initialization and has scalability issues. Kruskall's stress can be hard to interpret in terms of the quality of the dimensionality reduction techniques. It is also designed for Non-Metric Multi-Dimensional Scaling  which focuses on preserving the rank order of distances rather than the exact distances. On the other hand, similarity metrics other than distance metrics for tabular data, are limited to feature-wise or pair-wise metrics and few capture the overall structure of the datasets \cite{Dankar2022, Hernandez2022}.

\section{Background}\label{sec_back}
PCA is based on the idea of linearly transforming data into a new orthonormal basis where the principal directions capture the variation of the data. Each principal component is a linear combination of contributions from each original variable, and transforming data in accordance is interpreted as a projection along the principal component. 
In noisy data, PCA has the effect of concentrating the signal into the first few principal components, which can be useful for dimensionality reduction and data analysis \cite{Tipping1999, Jolliffe2002} by transforming data into a space where the elements are mutually uncorrelated.

Even though PCA is a well-known and established method, we discuss its mathematical foundation and nomenclature as a basis for the subsequent definitions of our new metrics.
Let the dataset $\bm X$ be an $(n\times d)$ matrix consisting of $n$ independent observations $\bm x_1,\bm x_2,...,\bm x_n$ in $d$ variables (i.e., ${\bm x_i \in \mathbb{R}^d}, {i=1,2,...,n}$). Translating to the mean-subtracted (i.e., centered) dataset $\Tilde{\bm X}$ with the same shape and elements $\Tilde{\bm x}_i = \bm x_i-\Bar{\bm x}$ with $\bar{\bm x}= 1/n \sum_i^n \bm x_i$; allows expressing the sample covariance matrix $\bm S$ by
\begin{equation}\label{eq:cov_s}
    \bm S = \frac{1}{n-1} \Tilde{\bm X}^T\Tilde{\bm X}.
\end{equation}
The first principal component scores are calculated as a linear combination of the original variables ${z_{i1}= \bm a_1^T \Tilde{\bm x}_i}$, with the vector of coefficients $\bm a_1$ to be found by maximizing the sample variance of the new variable:
\begin{align}
    \Var(z_{11},z_{21},...,z_{n1}) &= \frac{1}{n-1} \sum_{i=1}^n (z_{i1}-\Bar{z}_1)^T(z_{i1}-\Bar{z}_1) \nonumber \\
    &= \bm{a}_1^T \bm{S} \bm{a}_1,
\end{align}
subject to the constraint $\bm a_1^T\bm a_1=1$ to ensure a finite $\bm a_1$. This problem can be formulated as a Lagrange function:
\begin{equation}
    \mathcal{L}(\bm a_1, \lambda_1) = \bm a_1^T \bm S \bm a_1 - \lambda_1(\bm a_1^T\bm a_1-1),
\end{equation}
where $\lambda_1$ is a Lagrangian multiplier. The stationary solutions are found by taking the partial derivative w.r.t.\ $\bm a_1$,
\begin{equation}
    \frac{\partial}{\partial \bm a_1}\bm a_1^T \bm S \bm a_1 - \lambda_1 \frac{\partial}{\partial \bm a_1}\left(\bm a_1^T\bm a_1-1\right) = 0,
\end{equation}
which yields the eigenvalue equation $\bm S \bm a_1 = \lambda_1 \bm a_1$. Thus the variation is maximized when the vector of coefficients $\bm a_1$ is the eigenvector of the sample covariance matrix corresponding to the largest eigenvalue:
\begin{equation}
    \Var(z_{11},z_{21},...,z_{n1}) = \bm a_1^T \bm S \bm a_1 = \lambda_1.
\end{equation}
Scores for the second principal component ${z_{i2}=\bm a_2^T\bm x_i,}\: {i=1,2,...,n}$ are found similarly by maximizing the variance ${\Var(z_{12},z_{22},...,z_{n2})}$ due to $\bm a_2$ with the constraint $\bm a_2^T\bm a_2=1$, but with the addition that $z_{i2}$s be uncorrelated to~$z_{i1}$s. The optimum choice for coefficients $\bm a_2$ is revealed to be the eigenvector of the sample covariance matrix with the second largest eigenvector $\lambda_2$. Subsequent component scores are found by iteration. In general the optimal linear projection of the centered dataset $\Tilde{\bm X}$ along $p\leq d$ components is obtained by the transformation: 
\begin{equation}
    \bm Z = \Tilde{\bm X}\bm A,
\end{equation}
where $\bm Z$ is an $(n \times p)$ matrix of projection scores and $\bm A$ is a $(d\times p)$ matrix consisting of eigenvectors ${\bm a_1,\bm a_2,...,\bm a_p}$ of the sample covariance matrix corresponding to the $p$ largest eigenvalues. 

\section{Novel Metrics for Inter-Dataset Similarity} \label{sec_prm}
A primary function of PCA is to be a consistent way of representing data by transforming along the dimensions of maximal variance. Since the process is based on the empirical covariance matrix, it follows that two datasets that share the same fundamental statistics (or are a fair sample from the same distribution) will have a similar representation using PCA. In other words, if two datasets $\bm X, \bm X'$ share the same underlying statistics, then the difference between their PCA projections will be insignificant.
With this motivation and with the foundations established, we define two new metrics to capture the \emph{difference in explained variance} and the \emph{angle difference} between the first principal components.

\subsection{Difference in Explained Variance}
The first metric we present is the difference in variance explained by the principal components, i.e., the difference between the eigenvalues. The normalization factor is calculated based on $p$ and $d$. 
\begin{equation}\label{eq:vardiff}
    \Delta \lambda = \frac{d}{d+p-2} \sum_i^p |\lambda_i-\lambda_i'|
\end{equation}

The normalization factor in the explained variance metric (Eq.~\ref{eq:vardiff}) is subject to a few considerations. While metrics comparing absolute differences in lists of probability-like values, e.g., Total Variation Distance~\cite{Gibbs2002}, usually incur a normalization factor of 1/2 for constraining to the unit interval. However, in the comparison of explained variance (eigenvalues) difference, ordering of the eigenvalues ensures that the distributions can never be pure opposites. The amount of variance explained by the first principal component is always larger than that of the next principal component, hence, the maximal difference is when one dataset is entirely explained by the first component i.e., $\lambda_1\approx1, \lambda_2\approx0,...,\lambda_p\approx0$ and the other is pure noise $\lambda_1'\approx \frac{1}{d}, \lambda_2'\approx\frac{1}{d},...,\lambda_p'\approx\frac{1}{d}$. The maximal absolute difference is then: 
\begin{equation}
    \left|1-\frac{1}{d}\right| + \sum_i^{p-1} \frac{1}{d}= \frac{d+p-2}{d}.
\end{equation}
This value quickly approaches 2 for high-dimensional projection spaces, justifying this as a standard choice for datasets with many variables. When considering only a few principal component dimensions, e.g.~1, this choice will give value that is not properly constrained to the unit interval.   
 
\subsection{Angle Difference}
\begin{figure*}[tb!]
    \centering
    \includegraphics[width=\linewidth]{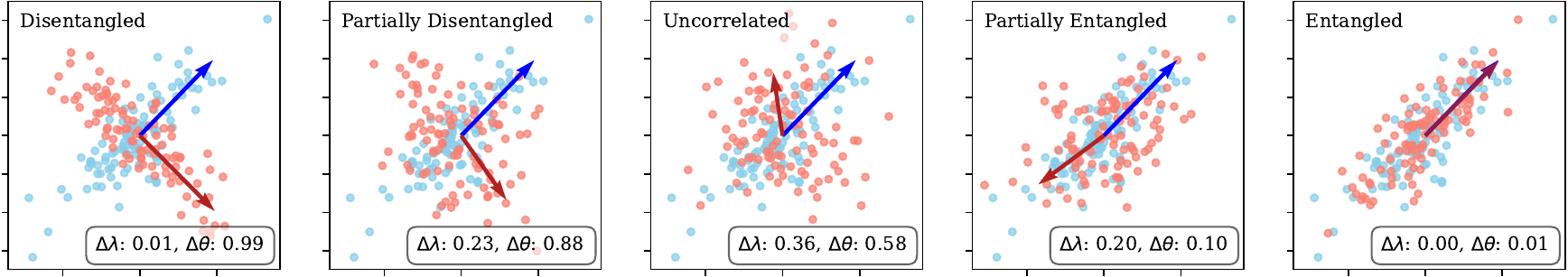}
    \caption{\textbf{Samples from Increasingly Similar Populations.} The figure shows a source distribution (in blue) and a query distribution (in red). From left to right, the query distribution is moved closer to the source distribution, changing the values of the proposed metrics $\Delta\lambda$ and $\Delta\theta$.}
    \label{fig:pca_illu}
\end{figure*}

The second metric measures the difference in the directions of the first principal components by calculating the angle between them. This describes if the two compared populations vary along different axes:
\begin{equation}\label{eq:anglediff}
    \Delta \theta = \frac{2}{\pi}\min \left[\arccos{(\bm a_1\!\cdot\! \bm a_1')}, \arccos{(\bm a_1\!\cdot\!(-\bm a_1'))}\right]
\end{equation}
We take the minimum of the angles, as the first principal components can be anti-parallel (see Fig.~\ref{fig:pca_illu}). We exploit several properties of the orthonormal basis in this calculation: We only need to check the first principal components as the remaining are orthogonal. Additionally, we drop the denominator term in the calculation of the angle since principal components are normalized. Finally, the metric is constrained to the unit interval by the normalization factor of $2/\pi$.

\begin{figure*}[tb!]
    \centering
    \includegraphics[width=0.99\textwidth]{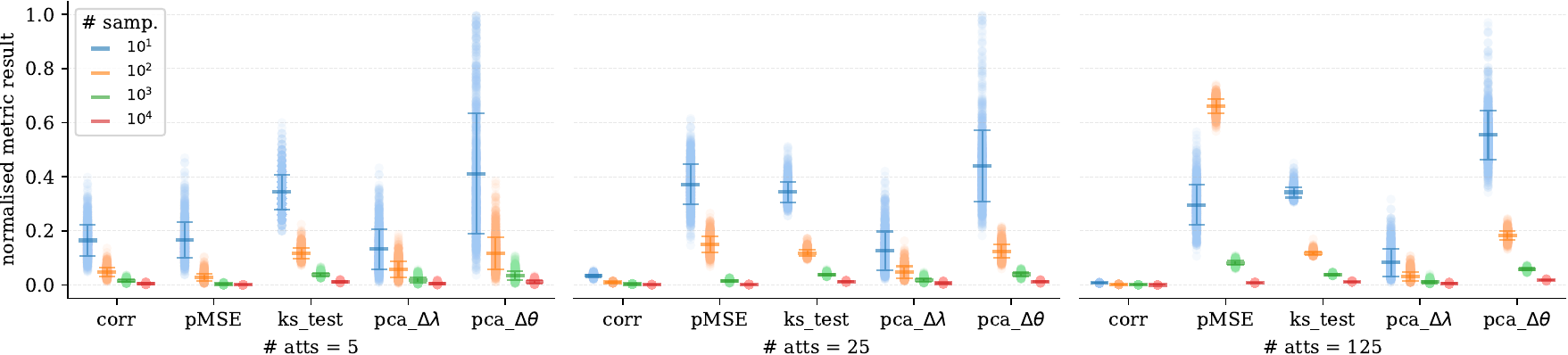}
    \caption{\textbf{Metric Variation with Varying Sample Size and Number of Attributes.} The proposed metrics and other state-of-the-art metrics (correlation matrix difference, propensity mean squared error, and the Kolmogorov-Smirnov test) do not conform to the optimum value, due to noise for fewer samples and sparsity from additional dimensions even when the compared samples stem from the same multivariate distribution.}
    \label{fig:metrics_std_sample_size}
\end{figure*}

\subsection{Instability to Sample Size}
We illustrate that two samples drawn from the same distribution will have a variance in both metrics if the sample size is small (see rightmost groups in Fig.~\ref{fig:metrics_std_sample_size}). Hence, these metrics are not expected to converge to zero for similar samples unless the sample size is sufficiently large. This also applies to other measures of similarity as exemplified in the same figure, where correlation matrix difference coefficient, propensity Mean Squared Error (pMSE), and the Kolmogorov-Smirnov test (see, e.g.,~\cite{Dankar2022}) all show the same instability for tens and hundreds of samples.

 

\section{Application in Synthetic Data Evaluation}\label{sec_synapp}
Synthetic data generation is a method for oversampling, augmenting, or protecting sensitive information in real-world datasets~\cite{Breugel2021, Yoon2020}. A high-quality synthetic dataset accurately mimics the underlying statistics of its non-synthetic counterpart but with no transferable samples. Evaluation of synthetic data is an active field of research with many open challenges. Several metrics exist for comparing synthetic and real data, but many focus on the marginal or pairwise similarity~\cite{Dankar2022, Hernandez2022}. In this section, we use the new metrics described above to measure multivariate similarity across two modes of comparison. The proposed metrics represent the global statistical alignment of the datasets under comparison and provide valuable insights towards overall statistical validity for tabular data that is primarily numerical. In principle, faithful synthetic data should have the same principal component representation as the real data. Thus the metrics we propose should not be more different than the expected deviation due to precision in estimating the empirical covariance matrix. This limitation also influences other popular metrics, as shown in Fig.~\ref{fig:metrics_std_sample_size}.

\begin{figure*}[tb!]
    \centering    \includegraphics[scale=0.688]{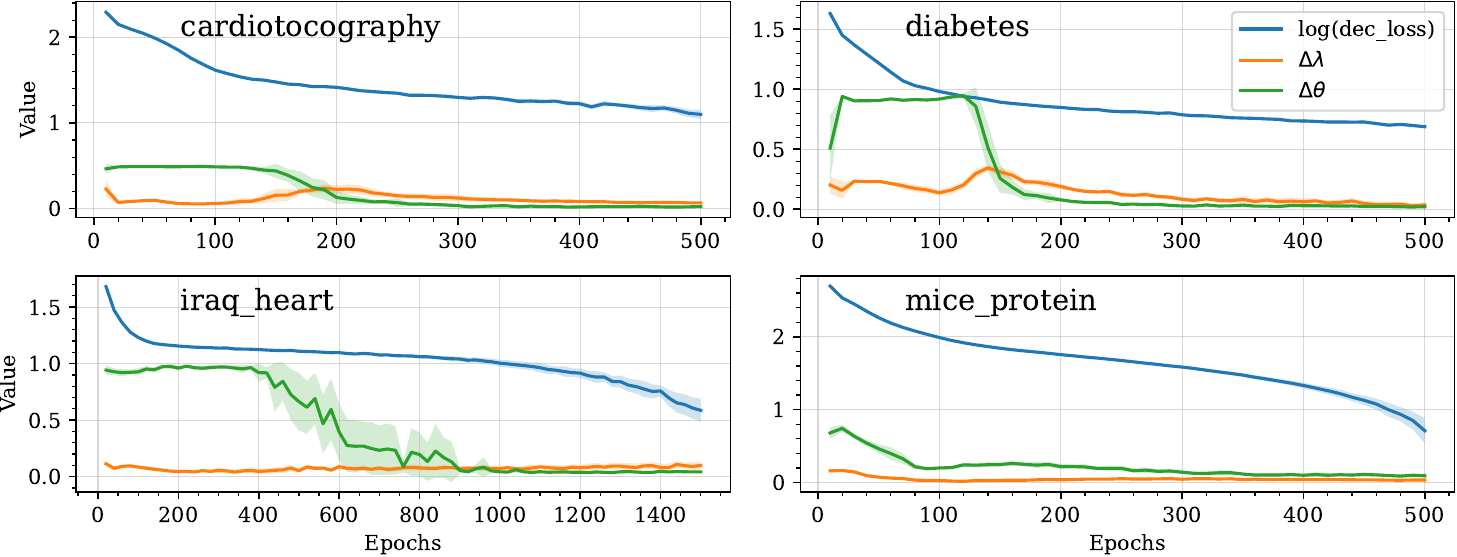}
    \caption{\textbf{TVAE Training Loss Alongside the Proposed Metrics.} After an initial period (the model learns characteristics of the data), the metrics drop, indicating a new alignment of the real and synthetic data. Afterwards the metrics steadily improve, aligned with the decreased loss. This demonstrates that the metrics are meaningful for measuring synthetic data quality. Confidence interval is 95\%, based on 10 independent runs.}
    \label{fig:tvae_loss}
\end{figure*}

An integral part of proposing new metrics for a certain task is to verify that the metric is useful for measuring the target phenomenon. As illustrated above, moving from disentangled distributions towards more similar distributions is the principal way of simultaneously improving the values of both metrics. Hence, improving the quality of synthetic data, e.g., during training, should improve the readouts of the two new metrics as training is progressing. In Fig.~\ref{fig:tvae_loss}, we show that this is the case for data produced by Tabular Variational Autoencoder (TVAE)~\cite{Xu2019} modelled on common datasets.

\begin{figure*}[tb!]
    \centering
    \includegraphics[width=0.99\textwidth]{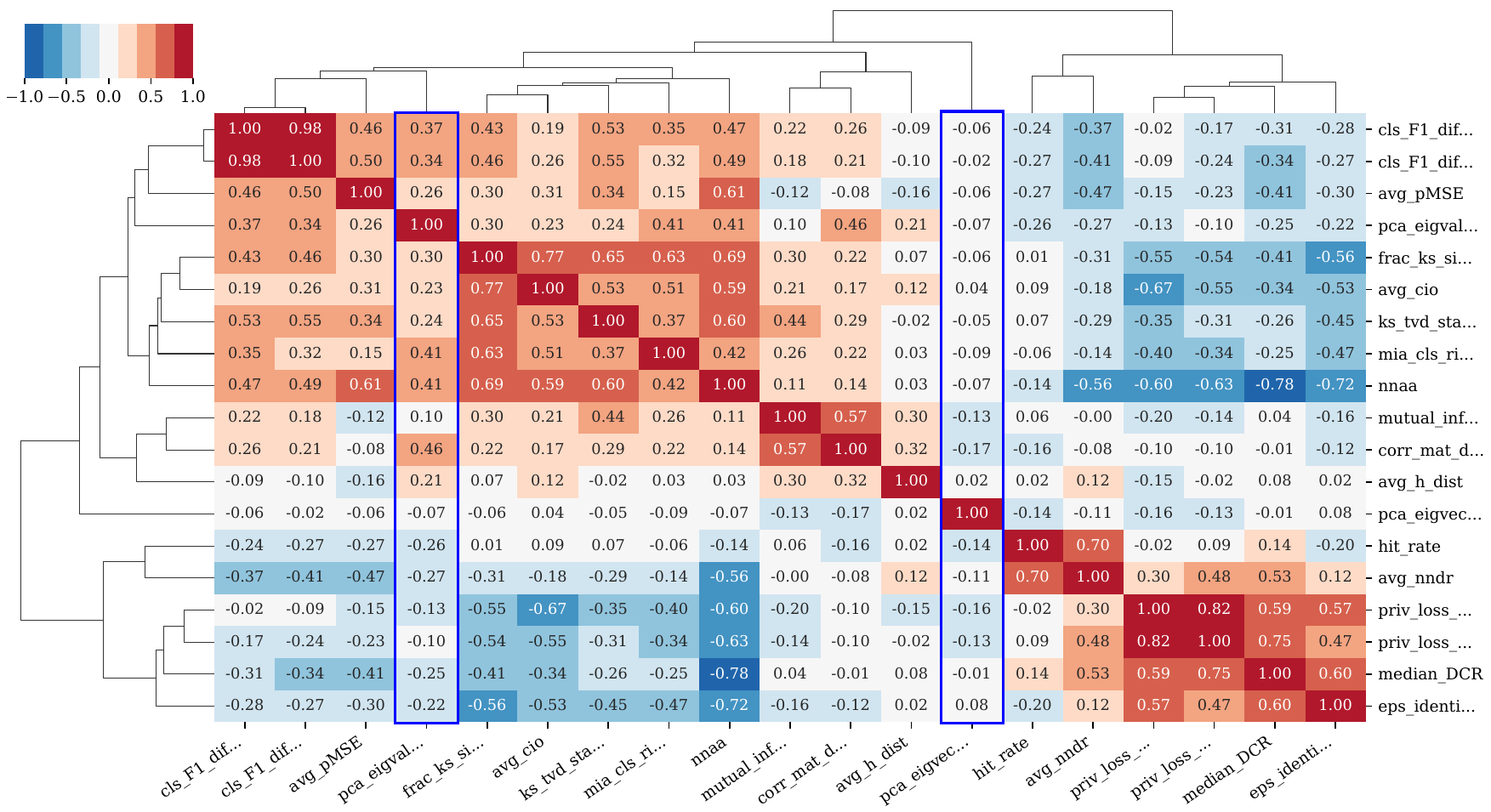}
    \caption{\textbf{Correlation Hierarchy of Metrics}. The heat-map is produced by taking the correlations of results of the metrics on 64 synthetic datasets made using different generative models. Closely associated metrics have little value in the same benchmarks since they describe the same mode of similarity/privacy leakage. The values of each metric were normalized such that a high value is better. The proposed metrics are outlined in blue.}
    \label{fig:corr_hier}
\end{figure*}

\subsection{Metrics Correlations}
To identify what these new metrics bring to the table, we evaluate them alongside a broad range of common metrics (using SynthEval~\cite{Lautrup2024}). Looking at how the metrics correlate with each other reveals structures of potential redundancy~\cite{Dankar2022}. In Fig.~\ref{fig:corr_hier} we show the results where the correlation heat-map is structured by hierarchical similarity clustering of the metrics. A clear distinction is seen between privacy and utility metrics each occupying a section of the diagonal, correlating negatively with each other. The proposed metrics are outlined in blue. The correlation matrix hierarchy figure is the result of an experiment where a selection of common benchmark datasets (Table~\ref{tbl:syn_ds}) for synthetic data tasks was generated using TVAE~\cite{Xu2019}, ADS-GAN \cite{Yoon2020}, Datasynthesizer \cite{Ping2017}, and the Synthpop CART model~\cite{Nowok2016}. Totaling, 64 synthetic datasets, which were evaluated across a broad range of metrics using SynthEval \cite{Lautrup2024}.

\begin{table}[h]
\caption{\textbf{Synthetic Data Generation Datasets.} Overview of the datasets used in the experiments with synthetic data generation (top four are ordered as featured in the main text, the rest alphabetically).}\label{tbl:syn_ds}\vspace{2pt}
\centering
\resizebox{\columnwidth}{!}{
\renewcommand*{\arraystretch}{0.8}
\begin{tabular}{lcrr}
\toprule
Dataset Name & Source & \# of Instances & \# of Features \\
\midrule
cardiotocography & \href{https://www.openml.org/search?type=data\&sort=runs\&status=active\&id=1466}{OpenML} & 2128 & 36 \\
diabetes & \href{https://www.kaggle.com/datasets/uciml/pima-indians-diabetes-database}{kaggle} & 770 & 9 \\
iraq\_heart & \href{https://www.kaggle.com/datasets/sukhmandeepsinghbrar/heart-attack-dataset}{kaggle} & 1319 & 9 \\
mice\_protein & \href{https://archive.ics.uci.edu/ml/datasets/Mice+Protein+Expression}{UCI} & 554 & 80 \\
breast cancer & \href{https://www.kaggle.com/datasets/reihanenamdari/breast-cancer/}{kaggle} & 4024 & 16 \\
cervical cancer & \href{https://archive.ics.uci.edu/dataset/383/cervical+cancer+risk+factors}{UCI} & 858 & 34\\
dermatology & \href{https://archive.ics.uci.edu/dataset/33/dermatology}{UCI} & 366 & 35 \\
drybean & \href{https://archive.ics.uci.edu/dataset/602/dry+bean+dataset}{UCI} & 13611 & 16 \\
hepatitis & \href{https://archive.ics.uci.edu/dataset/503/hepatitis+c+virus+hcv+for+egyptian+patients}{UCI} & 1385 & 28 \\
kidney\_disease & \href{https://archive.ics.uci.edu/dataset/336/chronic+kidney+disease}{UCI} & 400 & 24 \\
penguins & \href{https://www.kaggle.com/datasets/parulpandey/palmer-archipelago-antarctica-penguin-data?select=penguins_size.csv}{kaggle} & 344 & 7 \\
space\_titanic & \href{https://www.kaggle.com/competitions/spaceship-titanic/data?select=train.csv}{kaggle} & 6923 & 11 \\
stroke & \href{https://www.kaggle.com/datasets/fedesoriano/stroke-prediction-dataset}{kaggle} & 4909 & 11 \\
titanic  & \href{https://www.kaggle.com/c/titanic}{kaggle} & 721 & 8 \\
wine & \href{https://archive.ics.uci.edu/ml/datasets/Wine+Quality}{UCI} & 4898 & 12 \\
yeast & \href{https://archive.ics.uci.edu/dataset/110/yeast}{UCI} & 1484 & 8 \\
\bottomrule
\end{tabular}}
\end{table}

Evidently, the proposed metrics show only weak relationships with others, demonstrating their potential for new insights. Moreover, the closest metrics in the hierarchy are empirical metrics such as F1 diff and pMSE. This makes the proposed metrics appealing in contrast as they are fully deterministic for any sample and independent of the downstream tasks. F1 difference metrics (train and test) measure the performance of a selection of classifiers trained on real and synthetic data, respectively, in predicting a specified class of a real-data holdout sample \cite{Rankin2020}. Propensity MSE is a metric where a model is trained to distinguish between real and synthetic data and evaluated on a validation set~\cite{Snoke2018}. Both of these processes are non-deterministic and implementation-dependent. 
Furthermore, the correlation matrix difference is also somewhat linked to $\Delta\lambda$.

\subsection{Example} In the following we show how the metrics behave in a benchmark setting together with other common metrics. The dataset used is the Cardiotocography dataset \cite{Campos2010}. The benchmark is performed using SynthEval~\cite{Lautrup2024}, and the datasets are generated using various libraries (CTGAN~\cite{Xu2019}, Synthcity \cite{Qian2023}, Datasynthesizer \cite{Ping2017}, and Synthpop~\cite{Nowok2016}). From CTGAN we use the TVAE model from before, which is prompted to generate a dataset before, during, after, and long after the drop seen in Fig.~\ref{fig:tvae_loss} (corresponding to epochs 100, 180, 300, and 500). From Synthcity we use the ADS-GAN model (\textit{Anonymization through Data Synthesis using Generative Adversarial Networks} \cite{Yoon2020}). From Datasynthesizer we use the stock Bayesian Network model, and from Synthpop we use the sequential Classification and Regression Trees (CART) model.

\begin{table*}[t]
    \setlength{\tabcolsep}{11pt}
    \centering
    \caption{\textbf{Example Evaluation Results.} In this table, the evaluation results of multiple metrics and synthetic datasets (based on the cardiotocography dataset) are shown. Results in bold are the best in the column, and results in italics are the second best. All are rounded to two significant figures.}\vspace{2pt}
    \label{tab:examples}
    \renewcommand*{\arraystretch}{0.8}
    \begin{tabular}{l|cccccccc}
        \toprule
         & \multicolumn{2}{|c|}{Proposed Metrics} & \multicolumn{3}{|c|}{utility} & \multicolumn{3}{|c}{privacy} \\
         dataset & $\Delta\lambda$ & $\Delta\theta$ & Corr. & MI & KS-dist & hit-rate & $\varepsilon$-risk & MIR \\
         \midrule
         noisy (10\%) & \textbf{0.00} & \textbf{0.00} & \textbf{0.092} & \textbf{0.17} & \textbf{0.0020} & 0.58 & 0.98 & \textbf{0.34} \\
         independent &  0.45 & 0.40 & 6.9 & 2.7 & \textit{0.013} & \textbf{0.00} & \textbf{0.0010} & 0.41 \\
         TVAE 100 & 0.040 & 0.48 & 4.4 & 10. & 0.21 & \textbf{0.00} & \textit{0.014} & 0.44 \\
         TVAE 180 & 0.21 & 0.34 & 3.4 & 5.4 & 0.14 & \textbf{0.00} & 0.038 & 0.50 \\
         TVAE 300 & 0.064 & 0.013 & 2.6 & 2.7 & 0.096 & \textbf{0.00} & 0.059 & 0.48 \\
         TVAE 500 & 0.064 & 0.035 & 2.2 & 2.2 & 0.084 & \textbf{0.00} & 0.086 & 0.45 \\
         ADSGAN & 0.12 & 0.0069 & 1.8 & 1.8 & 0.082 & \textbf{0.00} & 0.040 & 0.41 \\
         datasynth. & 0.0018 & 0.0039 & 2.6 & 1.6 & 0.081 & \textbf{0.00} & 0.044 & 0.51 \\
         synthpop & \textit{0.00022} & \textit{0.0030} & 1.2 & \textit{1.1} & 0.022 & \textit{0.0021} & 0.23 & \textbf{0.34} \\
         best (hold-out) & 0.0033 & 0.0041 & \textit{0.98} & 2.0 & 0.031 & 0.016 & 0.30 & \textit{0.36} \\
         \bottomrule
     \end{tabular}\vspace{2pt}
     
     \footnotesize{Abbreviations:  MI -- Mutual Information, KS -- Kolmogorov Smirnov, MIR -- Membership Inference Risk.} \vspace*{-5pt}
\end{table*}

To expand the example further, we add some artificial examples to the benchmark: firstly a copy of the training data where we added 10\% normally-distributed noise to each entry, secondly a version where we independently resampled the columns of the dataset. Finally, a ``best'' dataset as a holdout dataset taken from the original dataset prior to training. This dataset represents an ideal synthetic dataset. It provides us with an indication of overfitting while illustrating a shortcoming in using the training set to measure utility and privacy. Table~\ref{tab:examples} presents the results.
The noisy dataset is the best, but among more realistic attempts, the synthpop CART dataset is the most reasonable for minimizing the disparity between synthetic and training data. In contrast, the ``best'' dataset receives worse scores than many others. Note how even this idealized dataset underperforms on privacy metrics knowing that there are no leaked samples in the holdout split. For TVAE, we see improvements in most metrics as the training progresses, including the proposed metrics. However, from epoch 300 to 500, none of the metrics improve significantly, signalling a convergence in terms of data quality. This property of the proposed metrics shows promise for early stopping in optimization problems.

\subsection{Implications for Privacy} In the example above, few of the datasets maintain an acceptable level of privacy. While the proposed metrics are not privacy metrics, they may help in sanity-checking results. Namely, invariance properties of PCA (depending on preprocessing choices) allow the proposed metrics to safeguard against privacy breaches from malpractice or ignorance. A copy of the real data that has been perturbed by noise, translated, or scaled will not alter the proposed metrics. Additionally, if the dataset has been rotated, only $\Delta\theta$ will change and by that amount. Proofs and further details are supplied in Appendix~\ref{app:proof}. 

\section{Application in Feature Selection Evaluation}\label{sec_fsapp}
Feature selection is the task of selecting the most informative and important features which represent the data in the best way. It is done to reduce the dimensionality while improving or maintaining the performance of the downstream machine learning or data mining task. Feature selection can be done in supervised or unsupervised settings. In supervised feature selection, the correlation of each feature with the labels or the impact of each feature in the final prediction of a classifier is measured. In unsupervised feature selection, the most informative features are selected without using any labels.

\subsection{Evaluation of Feature Selection Methods}
The quality of a feature selection method is often assessed by the performance of a downstream machine learning or data mining task. After the process of feature selection, the selected features are used for carrying out a classification or clustering task and the performance of those tasks determines the quality of feature selection. These evaluation techniques are not dedicated to the feature selection and are connected to either classification or clustering. 

In recent years, some dedicated metrics have been proposed to assess the quality of a feature selection algorithm. The Baseline Fitness Improvement (BFI) \cite{DBLP:journals/algorithms/MostertME21} is a  metric for the evaluation of feature selection algorithms which reflects the quality of a feature selection algorithm by taking the performance of the downstream task and the amount of dimensionality reduction into account. Feature Selection Dynamic Evaluation Metric (FSDEM) \cite{rajabinasab2024} is a dynamic evaluation metric for feature selection which yields deeper insights into the overall performance of a feature selection algorithm by monitoring its performance with different number of selected features and in overall. FSDEM can incorporate any supervised or unsupervised performance measure to evaluate the quality of an algorithm throughout the feature selection process. It also assesses the stability of the feature selection algorithm with respect to the potential gain by the inclusion of more features.

Despite the provision of useful insights into the performance of feature selection algorithms, especially by recently proposed metrics, all of the evaluation techniques in the area to some extent depend on a downstream machine learning or data mining task and hence are model-dependent. This limits the provided insights to a certain task or model and is an obstacle in the evaluation of the nature of a feature selection algorithm and it poses difficulties in the fair comparison of the performance of feature selection algorithms. 

\begin{figure*}[tb!]
    \centering
    \includegraphics[width=0.65\textwidth]{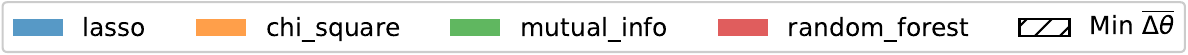}
    \includegraphics[width=\textwidth]{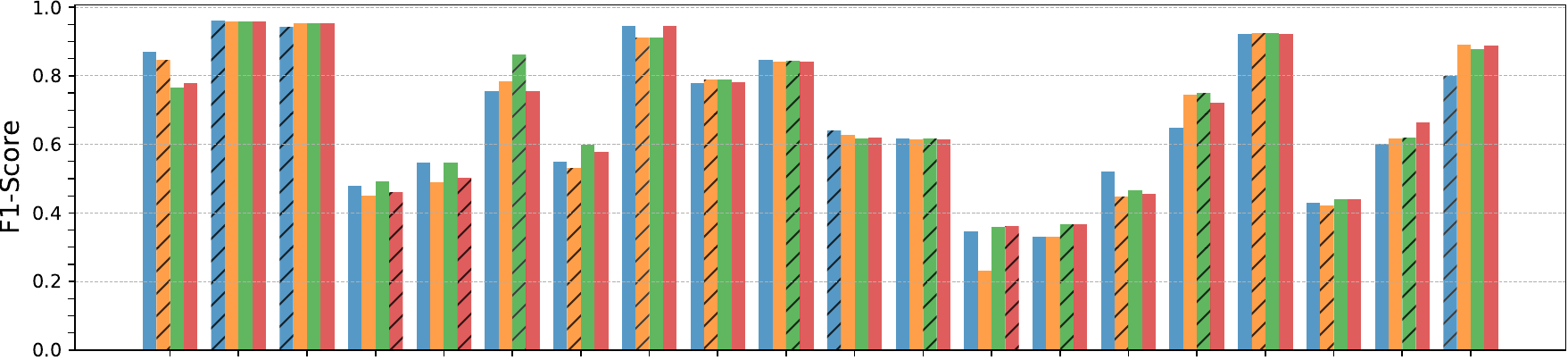}
    \includegraphics[width=\textwidth]{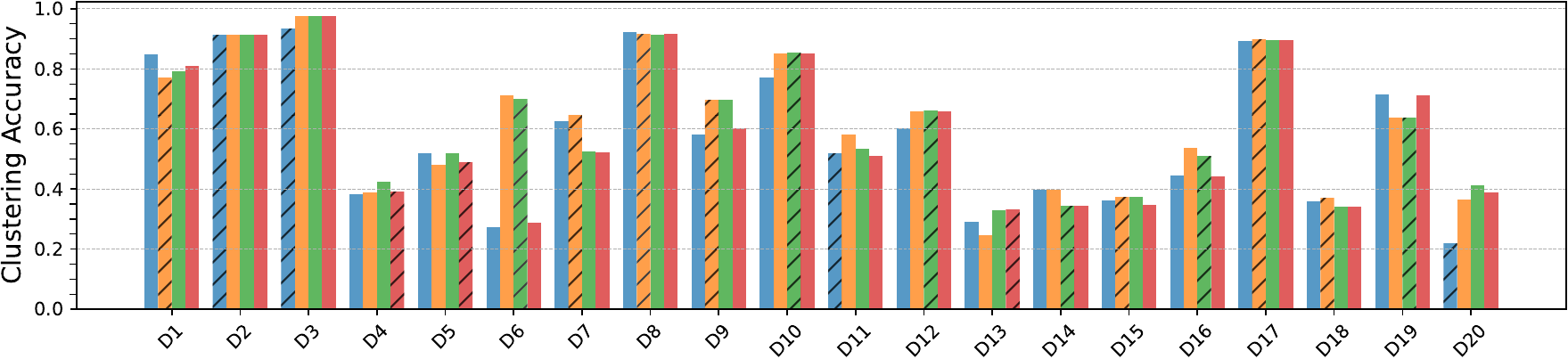}
    \caption{\textbf{Experimental Results for Feature selection.} The experimental results on 20 different datasets, 4 different feature selection methods and compared to two different model-dependent metric.}
    \label{fg:main_fs}
\end{figure*}

\subsection{Average Angle Difference}
We can utilize the angle difference property of the proposed metric as a model-agnostic approach for evaluating the performance of feature selection. Let $F$ be the set of features selected by a feature selection algorithm. Let $F^C$ be the complementary set of $F$ and $f$ a member of $F^C$. Let \textbf{$D_{f^0}$} be a copy of the dataset \textbf{$D$} where all instances of feature $f$ are equal to zero. We define Average Angle Difference (AAD), denoted by $\overline{\Delta\theta}$, as the average of the angle differences (Eq.~\ref{eq:anglediff}) of all \emph{not} selected features:
\begin{equation}
    \overline{\Delta\theta} = \frac{\sum_{f\in {F^C}} \Delta\theta(\textbf{D}, \textbf{D}_{f^0})}{|F^C|}
\end{equation}

AAD can be used for evaluating the quality of a feature selection algorithm without the need for using any downstream task or model. AAD yields a value in the range $[0, 1]$. 
We expect that removing redundant and non-informative features only contributes small angle differences to the measured average since they are more likely to have insignificant effect to the principal components.
As a result, smaller values of AAD are representative of a better feature selection process as they denote more similarity with the actual dataset after removing redundant features.

\subsection{Experimental Results}
In order to demonstrate the effectiveness of AAD for the evaluation of the performance of a feature selection algorithm, we conduct an empirical analysis on 20 different datasets from different domains. These datasets have different number of instances and features and they are all available publicly on UCI repository \cite{uci_ml_repository}. Table~\ref{apptbl:ds} shows the details of the datasets used in the experiments for feature selection evaluation in the paper. The datasets are ordered by the number of samples in the dataset. 

\begin{table}[tb!]
\caption{\textbf{Feature Selection Dataset Characteristics.} Overview of the datasets used in the experiments. These datasets all originate from the UCI Machine Learning Repository \cite{uci_ml_repository}.}\label{apptbl:ds}\vspace{2pt}
\centering
\resizebox{\columnwidth}{!}{
\renewcommand*{\arraystretch}{0.8}
\begin{tabular}{llrr}
\toprule
Key & Dataset Name & \# of Instances & \# of Features \\
\midrule
D1 & zoo & 101 & 16 \\
D2 & iris & 150 & 4 \\
D3 & wine & 178 & 13 \\
D4 & automobile & 205 & 25 \\
D5 & glass & 214 & 9 \\
D6 & dermatology & 366 & 34 \\
D7 & cylinder & 541 & 39 \\
D8 & breast cancer & 569 & 30 \\
D9 & credit & 690 & 15 \\
D10 & raisin & 900 & 7 \\
D11 & diabetic retino & 1151 & 19 \\
D12 & phishing & 1353 & 9 \\
D13 & yeast & 1484 & 8 \\
D14 & car eval & 1728 & 6 \\
D15 & steel & 1941 & 27 \\
D16 & cardiotocography & 2126 & 11 \\
D17 & rice & 3810 & 7 \\
D18 & nursery & 12960 & 8 \\
D19 & dry bean & 13611 & 16 \\
D20 & letter & 20000 & 16 \\
\bottomrule
\end{tabular}}\vspace*{-16pt}
\end{table}

In Fig.~\ref{fg:main_fs} we present the experimental results on the selected datasets. Four different feature selection methods, namely lasso \cite{tibshirani1996regression}, chi\_square \cite{liu1995chi2}, mutual\_info \cite{peng2005feature}, and a wrapper-based feature selection method \cite{kohavi1997wrappers}, are used for the evaluations. In order to show the effectiveness of AAD as a model-agnostic metric for evaluation of feature selection algorithms, we use the F1-score and clustering accuracy for comparison. The F1-score is calculated based on 10 different runs of 5-fold cross-validation for three different classifiers, namely decision tree, Support Vector Machine (SVM) with Radial Basis Function kernel, and k-Nearest Neighbor classifier (kNN) with $k = 5$. Clustering accuracy is calculated based on 10 different runs of $k$-means with number of clusters set to the number of classes. 

\begin{figure*}[tb!]
    \centering
    \includegraphics[width=\textwidth]{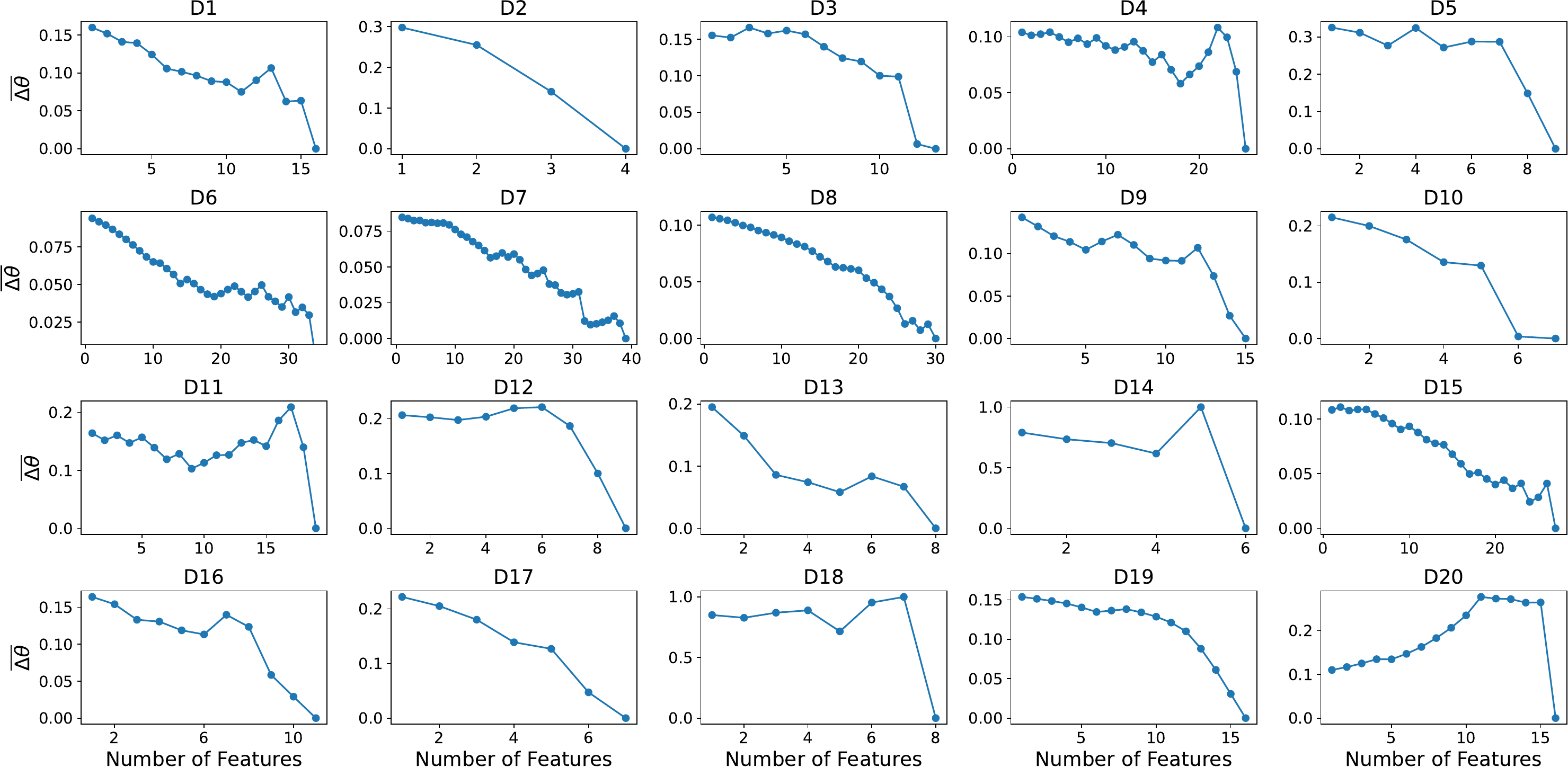}
    \caption{\textbf{Downwards Trend of Angle Difference} Results on 20 different datasets show that AAD decreases as more features are selected. Scale of AAD is adjusted to each dataset.}
    \label{fg:sec_fs}
\end{figure*}

The experimental results show that the minimum AAD value (indicated by a hatched bar) can successfully capture the best feature selection result, without depending on a machine learning model for the calculation of the metric. In most cases, the method identified by AAD is the best based on either F1-score or clustering accuracy (or both) and in other cases, it poses competitive figures. Note that, even though we tried to reduce the effect of models on the F1-score or clustering accuracy, they are still model-dependent. It is hard to determine the best overall feature selection result based on these metrics and we can only identify which feature selection result is best for classification or clustering using the selected classifiers for the experiments on average.

\begin{figure*}[tb!]
 \center
  \includegraphics[width=\textwidth]{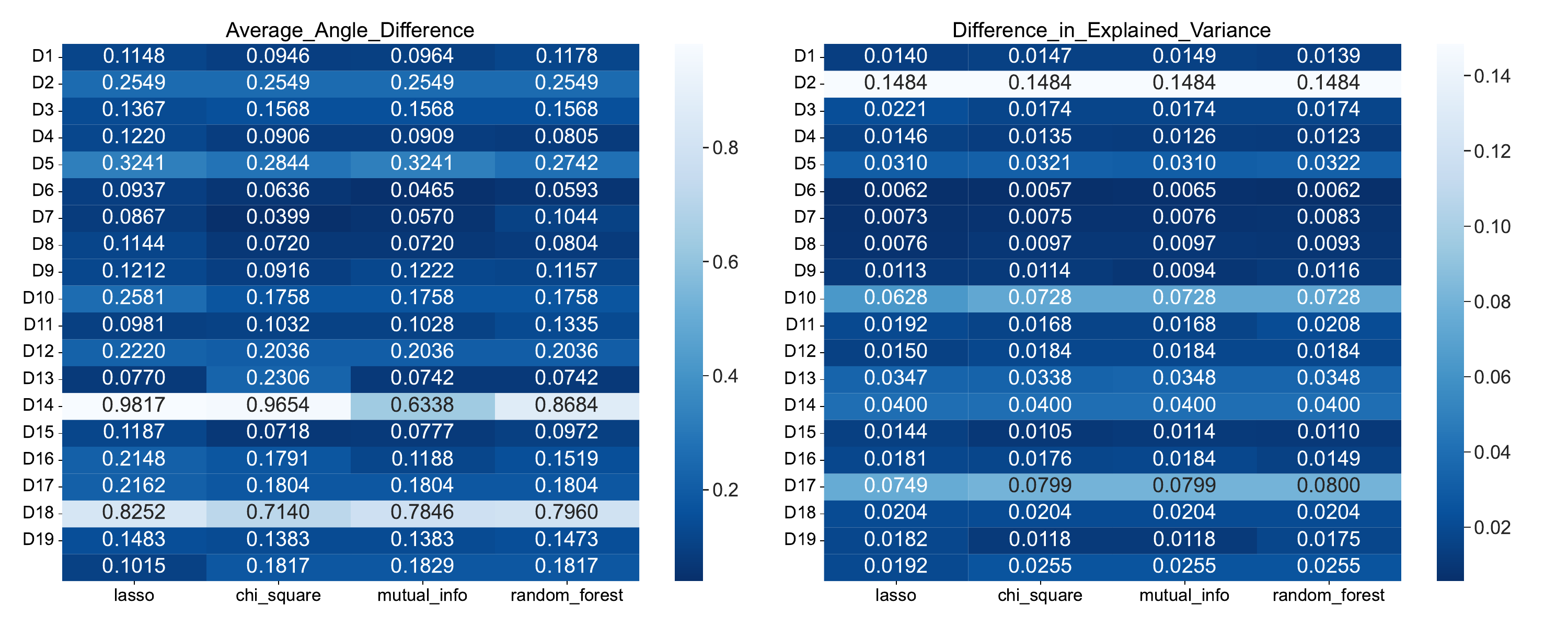}
  \includegraphics[width=\textwidth]{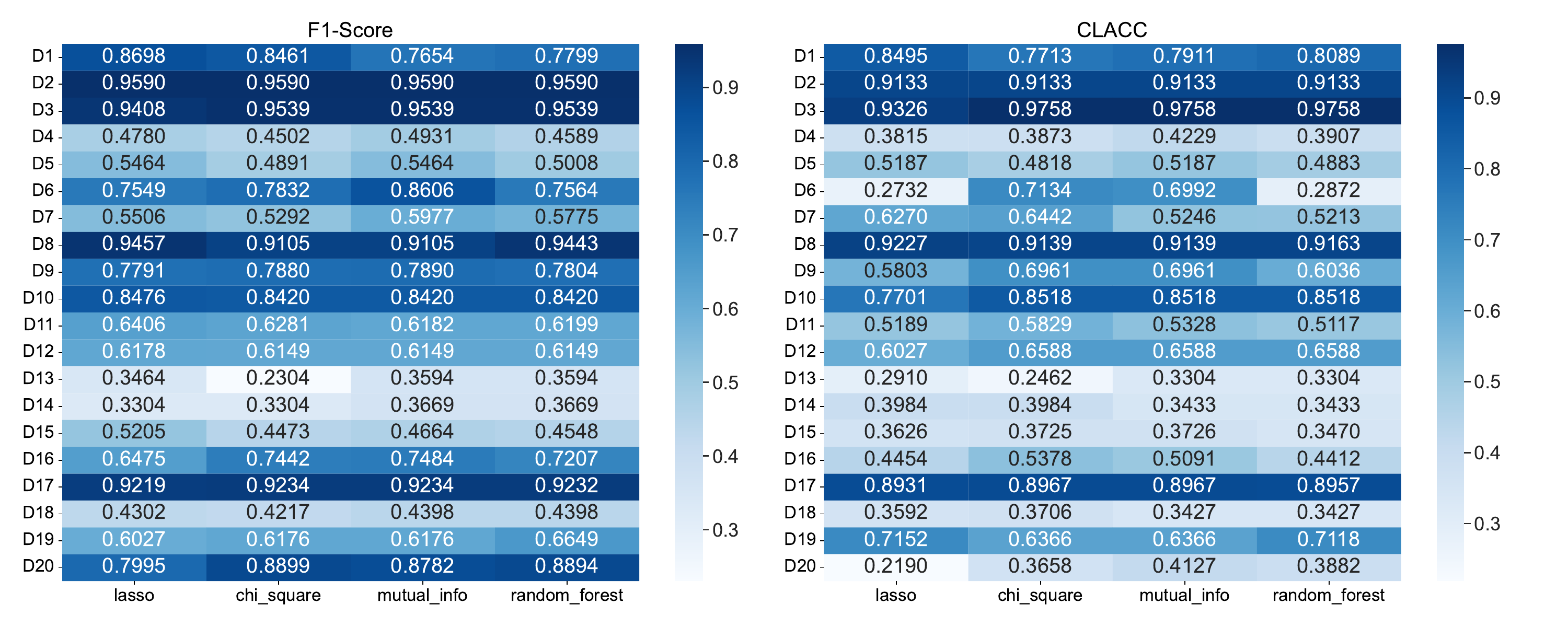}
  \caption{\textbf{Extended Experimental Results For Feature Selection} Results of experiments on 20 different datasets and for the proposed metrics, F1-score, and clustering accuracy. Color scale is reversed for the first two heat-maps as lower values are desired for those and higher values for the others.}
  \label{fg:app_fg1}
\end{figure*}

Extended experimental results for feature selection are presented in Fig.~\ref{fg:app_fg1}. This figure presents experimental results on the 20 different datasets and their respective values for AAD, difference in explained variance, F1-score, and clustering accuracy. Darker colors indicate better results for a metric (i.e., smaller values for AAD and difference in explained variance, larger values for the other metrics). We can see in the figure that there is an agreement in the smallest values of AAD for each dataset either with the highest F1-score or clustering accuracy. This demonstrates a use case for model selection: AAD can be used effectively as a model-agnostic approach to select the best feature selection result for each dataset by picking the result corresponding to the minimum AAD value across a dataset. In our experiments, there was an attempt to reduce the effect of the machine learning model on the results for other metrics by averaging through the performance of different classifiers, including SVM, kNN classifier and Decision Tree. However, it is noteworthy that these values are still model-dependent. This can be the reason behind the disagreements in the metrics.
These results along with the results presented in Fig.~\ref{fg:main_fs} demonstrate that AAD can be used as a model-agnostic approach for choosing the best feature selection algorithm.

\subsection{Empirical Analysis}
To further illustrate the effectiveness of AAD in capturing the quality of feature selection, we present the feature selection process in Fig.~\ref{fg:sec_fs}. In this figure, we show the change in AAD in the feature selection process on 20 different datasets selected for the experiments (note that the scale is adjusted for each dataset). We start with selecting one feature and we continue until all of the features in the dataset are selected. In most cases, the AAD value decreases as more features are selected, highlighting a clear downward trend. The fluctuations in the value of AAD can be due to the noise or the individual impact of a feature. Including  a feature individually may lead to an increase in the distance to the original dataset in the feature space in terms of principal components as it has an effect on the calculation. The inclusion of more features corrects this effect. Overall, there is a clear downward trend in the AAD values for most datasets which demonstrates the effectiveness of AAD for the evaluation of feature selection algorithms.

\section{Limitations and Future Work} \label{sec_lim}
The examples above illustrate the great potential for the proposed metrics. However, some limitations need to be addressed. One immediate limitation is that PCA is primarily defined for numerical data, however, real-world data often contain a mixture of numerical and categorical attributes. If a consistent ordinal encoding is used, PCA can  be applied to categorical attributes. Alternatively, if only a few columns are categorical, they can be omitted from the analysis and tested using different methods (e.g., Hellinger distance, Optimal Transport). Other types of data (e.g., image and text) can also be accommodated with taking additional pre-processing steps, as long as they are consistent. 

Another limitation relates to the implementation of the metric. Here, we use the same implementation throughout.
However, we observed that switching between using mean-subtracted data and data normalized by standard deviation affects the results. Even though this can be addressed by using a consistent normalization method, in future work, this effect should be systematically explored. Another limitation is that the metric values are not distributed perfectly. This problem raises complications in the interpretation of the metric values. In future work, we aim to make the distribution of practical measurements of the metrics fit into a more explainable range which could enhance the overall metric utility and transparency.

In this paper, we proposed two novel metrics for measuring inter-dataset similarity. We discussed their background and their mathematical foundation, and used two example applications in synthetic data and feature selection evaluation to thoroughly demonstrate their effectiveness in inter-dataset similarity measurement and its associated applications. We expect that further applications of the proposed metrics include other areas such as, ensemble learning, explainable AI, federated learning, transfer learning, dataset selection, or subspace quality assessment.\\

\noindent\textbf{Acknowledgement}\\
\vspace{-6pt}

\noindent This study was funded by Innovation Fund Denmark in the project “\small\textsc{PREPARE: Personalized Risk Estimation and Prevention of Cardiovascular Disease”}.



\printbibliography
\clearpage

\input{appendix}
\end{document}

%% file: appendix.tex
\appendix
\appendixpage
\label{app:proof}
In the appendix, we outline the mathematical results alluded to in the subsection on privacy implications. The section covered some examples of ``synthetic'' datasets where common privacy metrics would suggest acceptable privacy while the data in reality were just an altered copy of the original data. The~PCA metric was claimed to offer insights which may help avert such problems. 

\subsection*{Copied data} The first example is the trivial -- if the data are the same, then $\Delta\lambda$ and $\Delta\theta$ will indeed also be zero. Moreover, the privacy metrics are able to handle this case fine.

\subsection*{Noisy data} Recall the sample covariance element:
\begin{equation}\label{eq:cov}
    \Cov(\bm x, \bm y) = \frac{1}{n-1}\sum_i(x_{i} -\bar{x})(y_{i}-\bar{y}),
\end{equation}
with $\bar{x} = \frac{1}{n}\sum_i x_i$, and analogously for $\bar{y}$. Next, add normally distributed perturbation to every sample in every variable, such that $x_i\rightarrow x_i+\eta\delta x_i$, where $\eta$ is a scaling parameter and $\delta x_i \sim \mathcal{N}(0,\sigma)$. First, consider what happens to the averages;
\begin{equation}
    \bar{x} \to \frac{1}{n}\sum_i (x_i+\eta\delta x_i) = \frac{1}{n}\sum_i x_i +\frac{\eta}{n}\sum_i\delta x_i = \bar{x},
\end{equation}
the last term with $\delta x_i$ average to zero and thus we get the original mean. Inserting into Eq.~\ref{eq:cov},
\begin{multline}
    \Cov(\bm x, \bm y) \to \frac{1}{n-1}\sum_i(x_i+\eta\delta x_i - \bar{x})(y_i+\eta\delta y_i - \bar{y}) \\
    = \Cov(\bm x, \bm y) + \frac{\eta}{n-1} \sum_i\big(x_i \delta y_i + y_i \delta x_i  \\
     - \bar{x}\delta y_i - \bar{y}\delta x_i + \eta\delta x_i\delta y_i\big) \\
    = \Cov(\bm x, \bm y),
\end{multline}
with all the extra terms disappearing, again, due to the distribution of the noise being centred on zero. Thus it has been shown that random noise does not change the elements of the covariance matrix, and thus the PCA metrics are zero under this change. 

\subsection*{Spatial translation and scaling} While the covariance elements used in most of the theoretical treatment of PCA are only invariant to translations, and not scaling by separate factors, in practice, data are often transformed to a standardized form (i.e., $x_i \to \frac{x_i - \bar{x}}{s_x}$), which extends the covariance element into the Pearson correlation coefficient instead. The average of the new coordinate is zero:
\begin{align}
    \begin{aligned}
        \bar{x} & \to \frac{1}{n}\sum_i \left(\frac{x_i - \bar{x}}{s_x}\right) = \frac{1}{ns_x}\sum_ix_i-\frac{1}{ns_x}\sum_i \bar{x} \\
        & = \frac{1}{s_x}\bar{x}-\frac{1}{ns_x}n\bar{x}=0,
    \end{aligned}
\end{align}
then,
\begin{align}
    \begin{aligned}
        \Cov(\bm x, \bm y) &\to \frac{1}{n-1}\sum_i\left(\frac{x_i - \bar{x}}{s_x}-0\right)\left(\frac{y_i - \bar{y}}{s_y}-0\right) \\
        &=\frac{1}{n-1}\frac{1}{s_x s_y}\sum_i (x_i -\bar{x})(y_i -\bar{y}) \\ 
        &= \frac{\sum_i (x_i -\bar{x})(y_i -\bar{y})}{\sqrt{\sum_i (x_i-\bar{x})^2}\sqrt{\sum_i (y_i-\bar{y})^2}}.
    \end{aligned}
\end{align}
The new expression of the covariance element is invariant to scaling and translation of variables like ${\bm x\to a + b\bm x}$ and ${\bm y \to c + d\bm y}$ where $a,b,c,d \in \mathbb{R}$, and $b,d > 0$,
\begin{equation}
    \bar{x} \to \frac{1}{n}\sum_i (a+ bx_i) = \frac{1}{n}\sum_i a +\frac{b}{n}\sum_i x_i = a+b\bar{x},
\end{equation}
and
\begin{multline}
    \Cov(\bm x, \bm y) \to \\
    \qquad \frac{\sum_i (a+bx_i-a-b\bar{x})(c+dy_i-c-d\bar{y})}{\sqrt{\sum_i (a+bx_i-a-b\bar{x})^2}\sqrt{\sum_i (c+dy_i-c-d\bar{y})^2}} \\
    = \frac{bd\sum_i (x_i-\bar{x})(y_i-\bar{y})}{\sqrt{b^2\sum_i (x_i-\bar{x})^2}\sqrt{d^2\sum_i (y_i-\bar{y})^2}} \\
    = \Cov(\bm x, \bm y).
\end{multline}
Thus, it is seen, that under standardization of the variables, the $\Delta\lambda$ and $\Delta\theta$ metrics are invariant to translation AND scaling. Without standardization, the factors $b$ and $d$ carry through in the last part so $\Cov(\bm x, \bm y) \to bd\cdot\Cov(\bm x, \bm y)$.

\subsection*{Invariance to orthogonal transformations (e.g. rotation)} Alternatively, if the data are not normalized, the PCA metrics are able to detect orthogonal transformations such as a rotation (also possible if the transformations are applied after normalization). Consider for example, the dataset $\bm X = \bm Y \bm Q$ which is related to another similarly sized dataset $\bm Y$ through an orthogonal transformation~$\bm Q$, with $\bm Q \bm Q^{-1}=\mathbb{I}$. The covariance matrix can be factorized like:
\begin{equation}
    \bm S_Y = \frac{1}{n-1} \bm Y^T \bm Y = \bm B \bm \Lambda_B \bm B^{-1}
\end{equation}
into a decomposition of the matrix $\bm B$ consisting of the eigenvectors of $\bm S_Y$ and a diagonal matrix $\Lambda_B$ with the corresponding eigenvalues. Consider the decomposition of the related dataset,
\begin{align}
\begin{aligned}
    \bm S_X &= \frac{1}{n-1} \bm X^T \bm X = \frac{1}{n-1} \bm Q^T \bm Y^T \bm Y \bm Q \\
        &=\bm Q^T \bm B \bm \Lambda_B \bm B^{-1} \bm Q. 
\end{aligned}
\end{align}
Clearly $\bm S_X \neq \bm S_Y$, however, identifying $\bm A = \bm Q^T \bm B$ and $\Lambda_A = \Lambda_B$ shows that the two share common eigenvalues and projection $\bm Z = \bm X \bm A = \bm Y \bm Q \bm Q^T \bm B = \bm Y \bm B$. 

Because $\bm A = \bm Q^T \bm B$, the eigenvectors (principal components) are related similarly, i.e., $\bm a_i = \bm Q^T \bm b_i$. If the orthogonal transformation is a rotation $\bm Q = \bm R(\theta)$ the two sets of principal components will differ by the angle of the rotation only, i.e., $\Delta\theta=\theta$.